\documentclass{article}
\usepackage[square,sort,comma,numbers]{natbib}
\usepackage[preprint]{neurips_2019}

\usepackage[utf8]{inputenc} % allow utf-8 input
\usepackage[T1]{fontenc}    % use 8-bit T1 fonts
\usepackage{hyperref}       % hyperlinks
\usepackage{url}            % simple URL typesetting
\usepackage{booktabs}       % professional-quality tables
\usepackage{amsfonts}       % blackboard math symbols
\usepackage{nicefrac}       % compact symbols for 1/2, etc.
\usepackage{microtype}      % microtypography
\usepackage{amsmath,bm}
\DeclareMathOperator{\FL}{FL}
\DeclareMathOperator{\LL}{L}
\usepackage{graphicx} %插入图片的宏包
\usepackage{float} %设置图片浮动位置的宏包
\usepackage{subfigure} %插入多图时用子图显示的宏包
\usepackage{amssymb}
\usepackage{caption}
\newcommand{\etal}{\textit{et al}.}

\newcommand{\eg}{\textit{e}.\textit{g}.}

\title{Improving Semantic Segmentation via Dilated Affinity}

\author{%
  Boxi Wu \\
  CAD\&CG State key laboratory\\
  Zhejiang University\\
  Hangzhou \\
  \texttt{wuboxi@zju.edu.cn} \\
  \And
  Shuai Zhao \\
  CAD\&CG State key laboratory\\
  Zhejiang University\\
  Hangzhou \\
  \texttt{zhaoshuaimcc@gmail.com} \\
  \AND
  Wenqing Chu \\
  CAD\&CG State key laboratory\\
  Zhejiang University\\
  Hangzhou \\
  \texttt{wqchu16@gmail.com} \\
  \And
  Zheng Yang \\
  FABU Tech\\
  Hangzhou \\
  \texttt{yangzheng@fabu.ai} \\
  \AND
  Dengcai \\
  CAD\&CG State key laboratory\\
  Zhejiang University\\
  Hangzhou \\
  \texttt{dengcai@cad.zju.edu.cn} \\
}

\begin{document}

\maketitle

\begin{abstract}
Introducing explicit constraints on the structural predictions has been an effective way to improve the performance of semantic segmentation models. Existing methods are mainly based on insufficient hand-crafted rules that only partially capture the image structure, and some methods can also suffer from the efficiency issue.
As a result, most of the state-of-the-art fully convolutional networks did not adopt these techniques.
In this work, we propose a simple, fast yet effective method that exploits structural information through direct supervision with minor additional expense. 
To be specific, our method explicitly requires the network to predict semantic segmentation as well as dilated affinity, which is a sparse version of pair-wise pixel affinity. 
The capability of telling the relationships between pixels are directly built into the model and enhance the quality of segmentation in two stages.
1) Joint training with dilated affinity can provide robust feature representations and thus lead to finer segmentation results. 2) The extra output of affinity information can be further utilized to refine the original segmentation with a fast propagation process. Consistent improvements are observed on various benchmark datasets when applying our framework to the existing state-of-the-art model. Codes will be released soon.
	
\end{abstract}

\section{Introduction}
\label{introduction}
\begin{figure}
	\centering
	\includegraphics[width=0.9\linewidth]{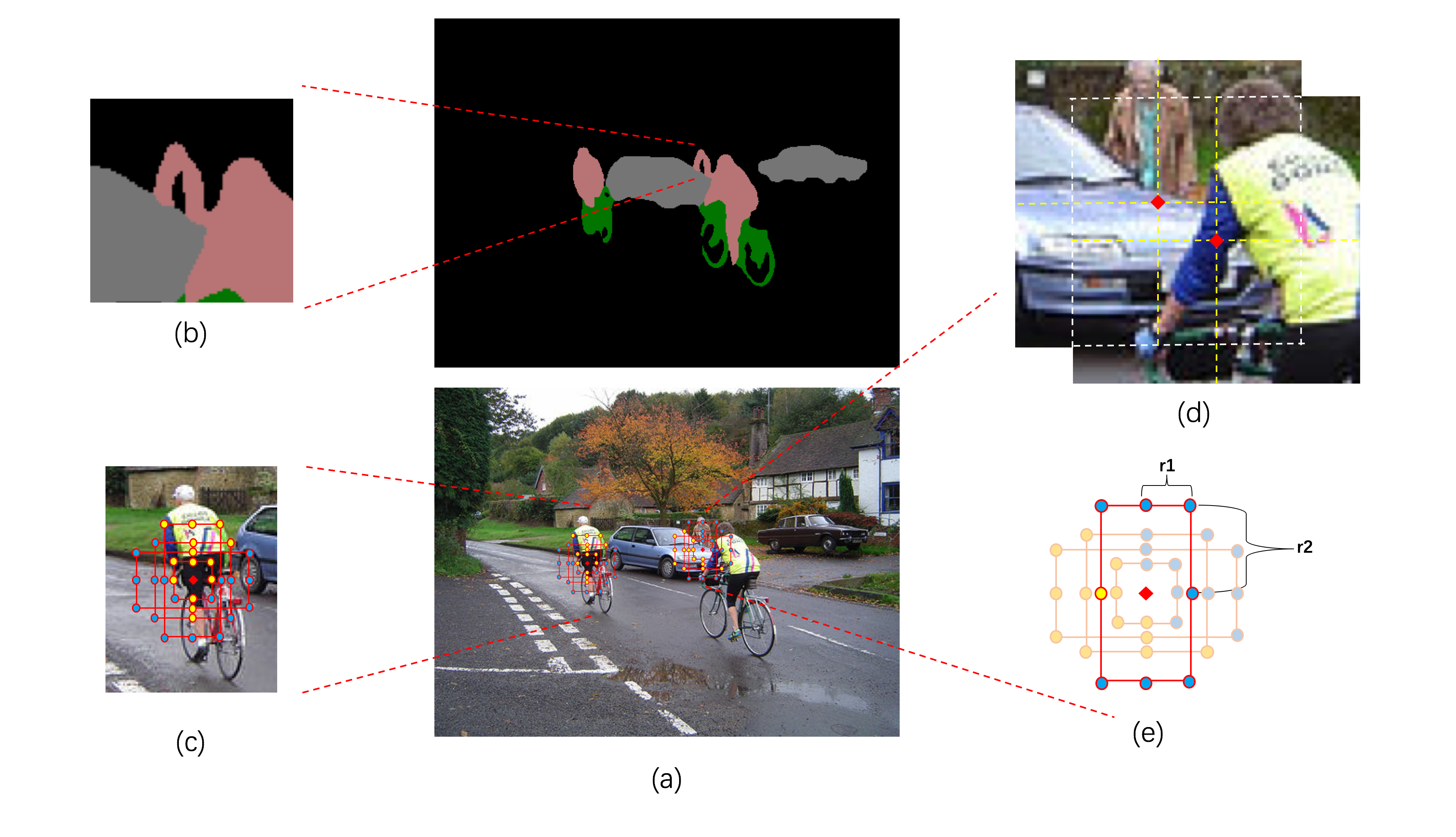}
	\caption{(a) The original image and the segmentation results of DeepLabv3+. (b) Noisy prediction can be found in the middle correct prediction. (c) Our dilated affinity provides structural information by predicting the similarity between pixels. Yellow points have the same class with the center point while blue ones belong to different classes. (d) Fully convolutional networks classify each pixel based on the image patch. Two similar patches with distinct labels are shown in the image. (e) Different rates of dilated affinity.}
	\label{fig:first}
\end{figure}

Semantic segmentation is one of the most fundamental and challenging computer vision problems. The goal of the task is to assign every pixel in an image a proper category label. With the breakthrough brought by deep learning, promising results have been achieved. Most of the cutting-edge methods belong to the fully convolutional networks~\cite{fcn}, which consider semantic segmentation as a pixel-wise classification problem.
To be specific, networks will derive the label of a single pixel solely from the image patch within its receptive field. 
The pixel and the corresponding patch become an independent sample, and relations between neighboring samples will be ignored other than they will be trained together in the same learning step.
This simplification can bring great challenges to classify pixels on the boundary area since those pixels may have distinct labels while their corresponding patches highly resemble each other, as shown in Fig.~\ref{fig:first} (d).
In addition, Fig.~\ref{fig:first} (b) shows that noisy predictions can often be found in the middle of correct predictions, which could have been avoided if the model also takes the surrounding predictions into consideration. 
Both failure cases can be alleviated if we have the structural knowledge of the relationships between pixels.

% former solutions disadvantage
Plenty of methods have been proposed to address this issue, including extra post-processing steps~\cite{crf,crfasrnn,mrf} and modified loss function\cite{aaf,unet,lmp}.
However, we find that these methods have various drawbacks. 
Most of them are manually designed based on heuristic priors, such as relations with raw pixel value~\cite{crf} or larger weights on boundary area~\cite{unet}. These priors are insufficient to fully capture the image structural information and can only deal with simple failure cases.
As the backbone network becomes stronger~\cite{deeplab2,deeplab3,deeplab3plus}, these solutions become less effective and may even lead to inferior results.
Besides, methods like \cite{crf,aaf} choose to model image structure on the final outputs of class distribution, where the rich information of high dimensional features~\cite{lrr} has been lost.
Post-processing methods like conditional random fields can also be time-consuming due to the dense modeling.
As a consequence, most of the leading models~\cite{deeplab3plus,exfuse,mcsi} have not utilized these aforementioned techniques.

% our method
Based on the analysis above, we propose to explicitly require the model to predict the similarity between pixels, which is the pair-wise pixel affinity information. To get a more efficient representation of image structure, we decide to model a sparse version of affinity and term it as dilated affinity.
The new task is parallel to semantic prediction, and thus it directly forces the high dimensinal feature maps to capture structural information during the training stage. 
This can be viewed as a substitution of customized loss functions.
Furthermore, we can use the dilated affinity information to refine segmentation with a fast affinity propagation post-processing. This extra step is inspired by the manual annotating process and can alleviate possible noisy predictions as well as vague edges.

There are several advantages of the proposed method.
First, our method is designed to fully utilize structural information based on direct supervision with minimum manual designed prior. Second, The moderate scale of dilated affinity is crucial to our success because dense modeling like CRF~\cite{crf} can lose locality and dramatically raise the computation cost. On the other hand, relations between adjacent pixels are overwhelmed with positive signals, leaving negligible information to explore. Dilated affinity avoids both of the problems.
Third, the ground truth of affinity information can be directly derived from the semantic label. This indicates that we do not ask for any extra information. Instead, we enhance the system based on what we have and try to explore every detail of existing information. Last but not least, it is easy to implement with different frameworks and require minor extra cost of computation.

% final part
We utilize the state-of-the-art DeepLabv3+~\cite{deeplab3plus} as our backbone networks. 
Notable improvements are observed on PASCAL VOC 2012~\cite{pascal_voc} and Cityscapes~\cite{cityscapes} datasets. Sec.~\ref{related_works} introduce related works of three directions, fully convolutional networks, image structural formation, and pixel affinity. Sec.~\ref{methods} explain our methods in detail and Sec.~\ref{experiments} show experimental results on PASCAL VOC 2012 and Cityscapes. Finally, in Sec.~\ref{conclusion} we conclude our framework.

\section{Related works}
\label{related_works}

\paragraph{Fully convolutional networks.} Fully convolutional network~\cite{fcn} is one of the pioneers that introduce deep learning into semantic segmentation and achieve impressive performance on benchmark datasets. Two important techniques are proposed and have been explored extensively afterward. First, they adapt networks that are originally designed for image recognition into a fully convolutional fashion and emit dense output directly. Later works found that dilated convolution~\cite{deeplab2}\cite{dilated} alleviates the precision decrease caused by excessive spatial destruction. In addition,  increasing the receptive field~\cite{parsenet,psp,deeplab3} can give extra performance boost. Second, skip architecture was proposed to refine the segmentation results with multiple level features and various substitutions~\cite{unet,lrr,deeplab3plus,exfuse,mcsi} have been investigated after that. 

\paragraph{Image structural information.} Works focusing on image structural information have also been developed. Ronneberger \etal~\cite{unet} decides to assign higher weights to samples on edges. Ke \etal~\cite{aaf} customizes the loss function to pull similar pixels together and push different ones away. Several post-processing methods choose to refine the predictions by aggregating the outputs on the image level. Conditional random field (CRF)~\cite{crf} is one of the earliest attempts on this direction, and many following methods try to enhance its capability, \eg, CRF as recurrent neural network~\cite{crfasrnn}, Markov random field~\cite{mrf} and spatial propagation~\cite{spn}. However, these attempts usually require much additional cost, both on time and memory, and fail to give rise to better performance when the backbone methods are sufficiently strong~\cite{deeplab3plus}\cite{exfuse}.

% add some paper about traditional affinity in adgm
\paragraph{Pixel affinity} Pair-wise pixel affinity is a fundamental computer vision concept and has been widely used under deep learning scenarios. Maire \etal~\cite{affinity-cnn} utilize affinity relation in the spectral embedding field while Liu \etal~\cite{spn} constructs a linear propagation module to learn pair-wise similarity matrix.
Recently, pixel affinity~\cite{adgm} and pixel link~\cite{pixel-link} innovatively model the problem as a task about telling whether two pixels belong to the same instance and have shown effectiveness on various practical scenes. 
We draw on their experience and modify the current state-of-the-art methods to enable the model to tell whether two adjacent pixels belong to the same class rather than the same instance.

\section{Methods}
\label{methods}

In this section, we first illustrate the concept of dilated affinity and explain several specific designs which allow it to fully cooperate with the original task of segmentation. Then we introduce the details of our network architecture and loss computation. In the end, we describe the affinity propagation post-processing which combines the coarse segmentation results and affinity information.

\subsection{Dilated pixel affinity}

Pair-wise pixel affinity is the concept which describes the similarity between pixels and may have different mathematical definition depending on the context of the problem. Under the scenario of semantic segmentation, we denote the affinity of two pixels with a binary signal and assign it with a positive value of 1 when the two pixels belong to the same class , otherwise 0. As shown in Eq.~\eqref{affinity_definition}, $y_1$ and $y_2$ are the semantic labels of pixel $x_1$ and $x_2$. And $a_{1,2}$ is their affinity:
\begin{equation}
a_{1,2}=a_{2,1} =
\begin{cases} 
1 &\text{if} \quad\!\! y_1=y_2 \\ 0 &\text{otherwise}
\end{cases}
\label{affinity_definition}
\end{equation}

When capturing the affinity of a pair of pixels, we only consider pixels within a restricted area since distant pixels lose locality and the complexity of modeling every pair of pixels grows rapidly with the size of the feature map. 
On the other hand, nearby pixels are also discarded as the signals are overwhelmed with positive values, leaving negligible information to explore. 
Thus, we decide to sparsely sample from pixels with reasonable distances in the same way of dilated convolution~\cite{dilated,deeplab2} and term this sampling method dilated affinity. As shown in Fig.~\ref{fig:first} (e).

To be specific, for pixel $x_{i,j}$ at position $(i,j)$ on the feature map, the network is required to predict its affinity with a group of pixels, which are on the 8 directions of dilation rate $r=(r_1,r_2)$. We denote this single group of pixels as $S_{r}(i,j)$.
Inspired by \cite{adgm,focalloss}, multiple groups of pixels with various dilation rates are taken into consideration. We denote the set of dilation rates as $R$ and the set of all targeted pixels as $S_{i,j}$. Their relations are shown below.

\begin{equation}
	S_{i,j} = \bigcup_{\bm{r}\in R} S_{r}(i,j),
\end{equation}
\begin{equation}
  S_{r}(i,j) = \{ x_{i+s,j+t}\big|~ \forall s\in \{-r_1,0,r_1\},~\forall t\in \{-r_2,0,r_2\} \setminus x_{i,j} \}.
\end{equation}

The influence of different choices of $R$ is explored in Sec.~\ref{experiments}.

\subsection{Architecture and loss computation}

\begin{figure}
	\centering
	\includegraphics[width=0.9\linewidth]{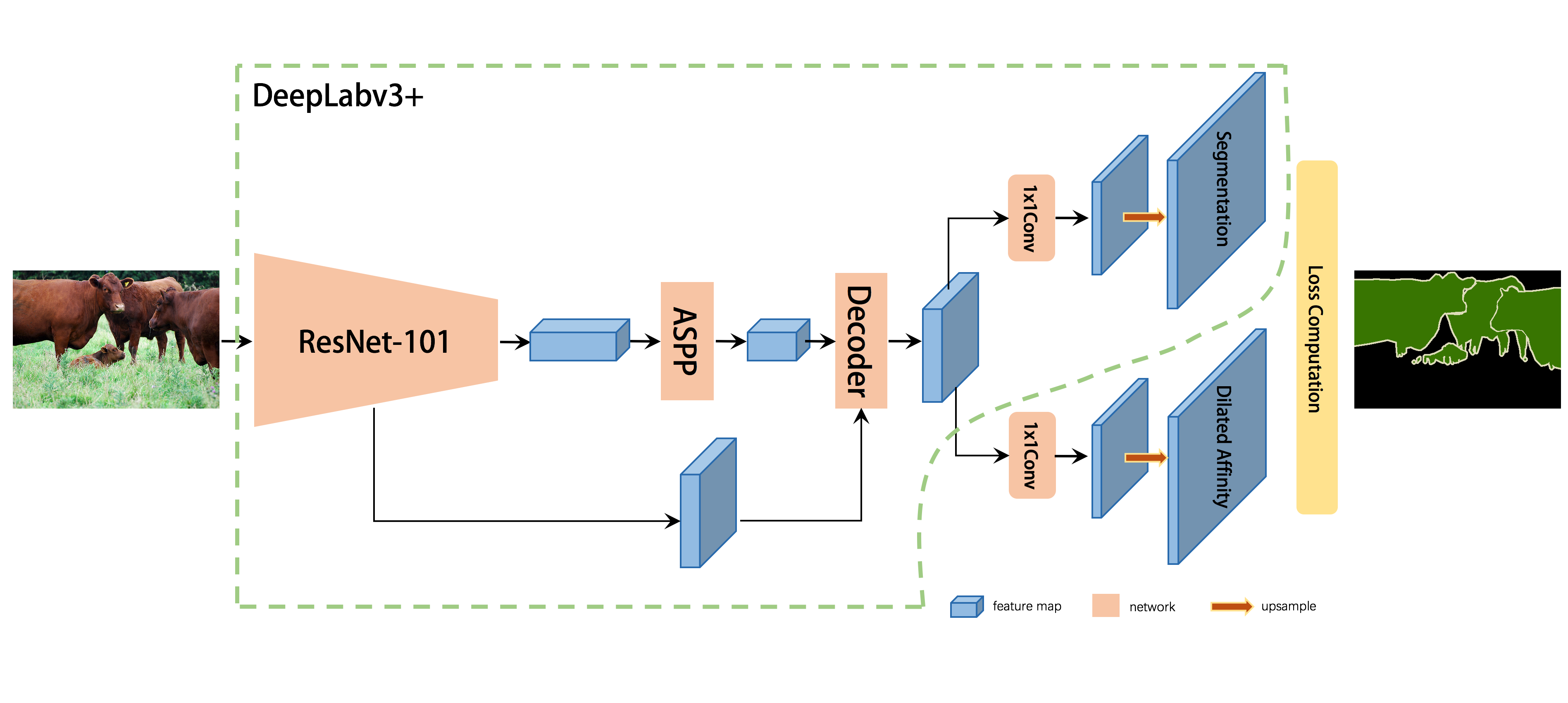}
	\caption{The overall architecture of our method.}
	\label{fig:architecture}
\end{figure}

% architecture
We choose the extraordinary DeepLabv3+~\cite{deeplab3plus} with ResNet-101~\cite{resnetv1} as our baseline model.
In the original DeepLabv3+, the feature map generated by ResNet-101 first go through the Atrous Spatial Pyramid Pooling (ASPP) module~\cite{deeplab2}. Then it is refined by a decoder, resulting in a 256-dimension feature map four times smaller than the original input. At last, a 1x1 convolutional layer is used to reduce the channel size to the number of classes and bilinear resize interpolation is used to upsample the feature map to the size of the original input. Softmax cross entropy function is adopted to compute the final loss.

To allow the network to predict dilated affinity, we add an extra 1x1 convolutional layer onto the feature map generated by the decoder. The new branch is parallel to the original segmentation branch. This parallel design is more sufficient to capture affinity information than \cite{aaf} since the similarity is derived from high dimensional features rather than low dimension probability vectors. The channel size of the branched outputs is equal to $8|R|$. Bilinear resize is still be used but the softmax operation is replaced with the sigmoid operation. The overall architecture is shown in Fig.~\ref{fig:architecture}.

% neighbor reweight
Dilated affinity learning resembles one-stage object detection in many ways that both of them are required to output extra information other than semantic predictions, and both of them face severe sample imbalance on the additional output. 
Dilated affinity with small rates still faces the problem of signal imbalance that positive signals occupy most of the proportion. 
To alleviate the signal imbalance talked above, we resort to focal loss~\cite{focalloss} and reweighting different samples.

\begin{align}
\FL(p_t) = -(1 - p_t)^\gamma \log(p_t) 
\quad \text{where} \ 
p_t=
\begin{cases} 
p & \text{if} \quad\!\! y_a=1 \\ 1-p & \text{otherwise}
\end{cases} 
\label{focal_definition}
\end{align}
The form of focal loss function is shown in Eq.~\eqref{focal_definition}, where $y_a\in\{0,1\}$ is the ground-truth of dilated affinity, $p\in[0,1]$ is the estimated probability of positive affinity and $\gamma\geq0$ is the \textit{focusing} parameter~\cite{focalloss}, which is set to 2 in our experiments.

\begin{figure}[h]
	\label{fig:nstat}
	\centering
	\subfigure[PASCAL VOC 2012 train set]{
		\includegraphics[width=0.4\linewidth]{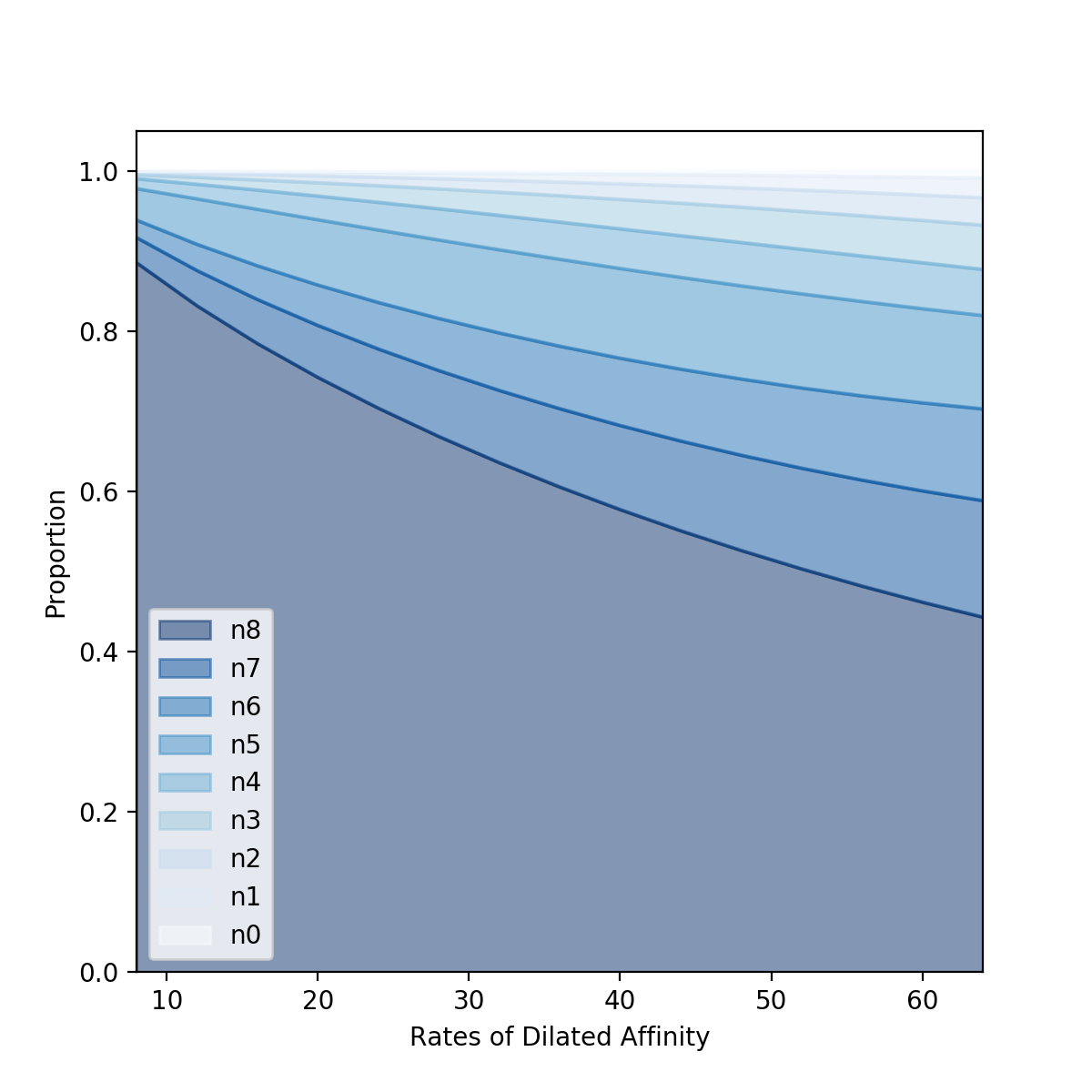}}
	\subfigure[Cityscapes train set]{
		\includegraphics[width=0.4\linewidth]{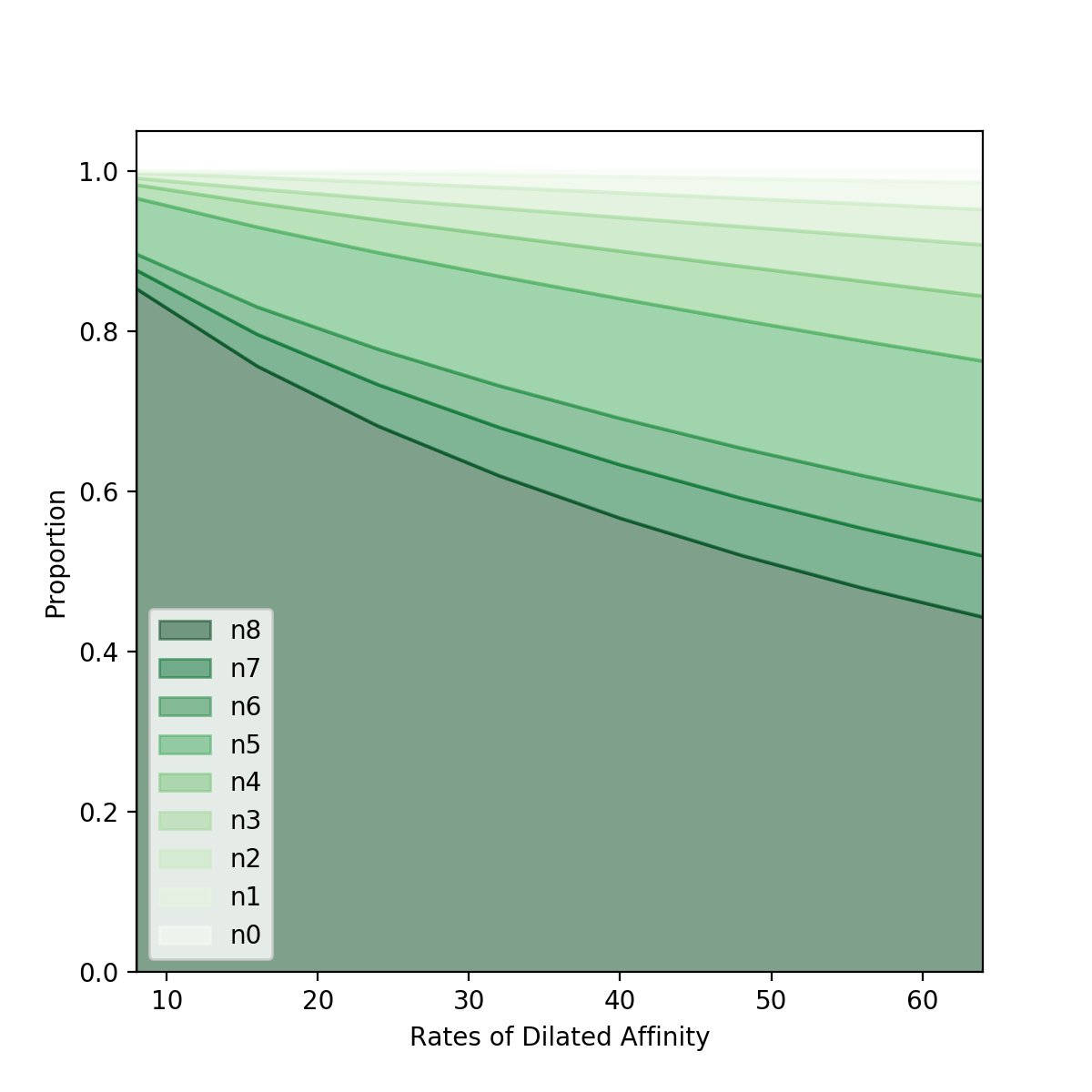}}
	\caption{The proportion of $n_0$ to $n_8$ changes with the rate of dilated affinity. Vertical axis shows the percentage of $n_0$ to $n_8$ with respect to all the pixels. Horizontal axis shows the corresponding rate. Image (a) and (b) is the statistics of PASCAL VOC 2012 train set and Cityscapes train set respectively.}
	\label{fig:neighbour}
\end{figure}

As for the weighting scheme, a straightforward approach would be reweighting different samples based on the inverse frequency of different signals~\cite{focalloss}, whether positive or negative.
However, it would be more appropriate to reweight based on independent samples rather than subdivided signals. And for a pixel in the boundary area, its positive affinity signals are as valuable as the negative ones.
Thus we opt for a different solution.
For pixels in set $S_{r}(i,j)$, we divide pixels into nine categories based on the number of positive affinity signals from their eight neighbors.
These categories are denoted as $n_0$ to $n_8$. We calculate the proportion of $n_0$ to $n_8$ for each $r \in R$, and use their inverse frequencies as weights instead.
The distribution of pixels for different datasets and different dilation rates are shown in Fig.~\ref{fig:neighbour}.

Directly using the inverse frequency based on positive neighbors may result in absurdly large weights to samples of $n_0$ and $n_1$ and experiments indicate that this aggressive weighting scheme can cause damage to the learning of semantic segmentation.
Thus we propose to use the square root value of inverse frequency instead. The final form of affinity loss is shown below.

\begin{equation}
\LL(\bm{p},\bm{y}_a) = \beta \sum_{r}^R         \frac{\mathnormal{freq}(n_8,r)}{\mathnormal{freq}(n_{\mathnormal{sum}(\bm{y}_a)},r)}       \sum_{c=1}^8\FL(p_t)
\label{total_pl}
\end{equation}

In Eq.~\eqref{total_pl}, $freq(n_k,r)$ is the frequency of $n_k$ for dilation rate $r$, and $sum(\bm{y}_a)$ is the amount of positive signal in $\bm{y}_a$.
As it suggests, the weight of pixels from $n_8$ is always equal to $1$.
The affinity loss is multiplied with a parameter $\beta$ before added to the total loss, and its value is selected via cross-validation. 
We find that it is crucial to assign $\beta$ a value large enough to gain improvements from the joint training procedure. However, too large $\beta$ can cause damage to the learning of semantic segmentation, although the accuracy of dilated affinity may keep rising.

\subsection{Refine segmentation with affinity propagation}

As discussed in Sec.~\ref{introduction}, It is more robust to classify a pixel not only based on the image patch, but also the neighboring pixels and the affinity information of them.
Also, when inspecting on the manual annotating process, we realize that humans do not classify every pixel separately since it is not only laborious but also unnecessary. 
What they actually do is recognizing the majority of the pixels, then refining the annotation by considering the similarity between them, especially on edges.

Inspired by this observation, we decide to refine the classification of a pixel by adding an additional factor, which is proportional to the predictions of nearby pixels and the corresponding affinity.
To fully express our intention, we define a general form of the refinement below and discuss the specific design in detail.

\begin{equation}
\hat{p}_{i,j} = \mathcal{N}(\ \lambda p_{i,j}\max(p_{i,j}) + \sum_{s\in S_{i,j}}a_sp_s \ )
\label{general_fg}
\end{equation}

In Eq.~\eqref{general_fg}, $p_{i,j}$ is the class prediction of pixel $x_{i,j}$, and $\hat{p}_{i,j}$ is the refined prediction. Both $p_{i,j}$ and $\hat{p}_{i,j}$ are vectors. $\lambda$ is a weight parameter whose value is selected by cross-validation.
For a pixel $s$ in $S_{i,j}$, we simplify its affinity as $a_s$, and its prediction of different categories as $p_s$.
$\mathcal{N}(\cdot)$ is the normalization function, which will make sure that the sum of $\hat{p}_{i,j}$ is equal to 1.
The max value of $p_{i,j}$ is utilized to keep confident predictions consistent with original predictions. Furthermore, because $a_s$ is always positive and may introduce noise when its value is small, we change the sigmoid operation of $a_s$ to a steeper version during the post-processing, as shown in Eq.~\eqref{steep_sigmoid}.

\begin{equation}
g(x) = \frac{\mu^x}{1+\mu^x} \quad \mu=7
\label{steep_sigmoid}
\end{equation}
This design can force small affinity value to zero and decrease the difference between high affinity signals.
It is noted that this refining process can be executed with multiple times like CRF~\cite{crf}, propagating the original classifications through the connection of positive affinity. Unlike the edge merge process~\cite{adgm}, our affinity propagation is faster and more robust since there is no demand to output instances results. The performance of different iteration times are shown in Sec.~\ref{experiments}

\section{Experiments}
\label{experiments}

\subsection{Experimental setup} 
\textbf{Base model.}
We choose the state-of-the-art DeepLabv3+~\cite{deeplab3plus} as our baseline.
The backbone network is ResNet-101~\cite{resnetv1},
which is only pretrained on ImageNet~\cite{imagenet}.
The parameters in the ASPP module and the decoder are randomly initialized.
Biases of the affinity layers are initialized as
$b_r = -\log(\pi_r/{1-\pi_r})$ following~\cite{focalloss}, where $\pi_r$ is set to the frequency of positive signals for $r\in R$.

\textbf{Experimental platform.}
All experiments are conducted on 4 GTX 1080Ti GPUs with Tensorflow~\cite{tensorflow}.
Furthermore, we implement the Cross-GPU Batch Normalization~\cite{megdet} 
to alleviate unstable statistics caused by small batch size on each GPU~\cite{megdet,gn}.

\textbf{Dataset.}
We evaluate our methods on two datasets, PASCAL VOC 2012~\cite{pascal_voc} and Cityscapes~\cite{cityscapes}.
PASCAL VOC 2012 is a semantic segmentation task that has 20 foreground classes and 1 background class. We utilize the extra annotation provided by~\cite{pascal_voc_aug}, resulting 10\,582, 1\,449 and 1\,456 for train, validation and test separately.
Cityscapes~\cite{cityscapes} is a large-scale semantic segmentation dataset, most of which are street scenes of various cities. We train the model on the 2975 images in the training set and test it on the 500 images in the validation set.

\textbf{Learning rate and training steps.}
For PASCAL VOC 2012, we train the model for $30K$ iterations with the crop size of 513 and the batch size of 16.
As for Cityscapes, the training steps is $90K$, crop size changes to 769, while the batch size decreases to 12.
The output stride, which is the spatial resolution ratio of the input image to the final output, is set to 16 during training. 
The rest settings, such as learning rate schedule and data augmentation are the same with DeepLabv3+~\cite{deeplab3plus}.

\textbf{Evaluation metric.}
The evalutation metric is the mean intersection-over-union (mIoU) score.
The output stride changes to 8 during evaluation.

\subsection{Ablation study on PASCAL VOC 2012}

We set $R$ to $\{8,16,32,64\}$ to compare different choices of weighting scheme and loss functions. The baseline uses equal weights and cross-entropy loss function. All the other three weighting schemes, \eg inverse frequency based on signals~(signal-reweight), inverse frequency based on neighbors~(neighbor-reweight), and the square root of the inverse frequency based on neighbors ~(sqrt-reweight) are adopting focal loss function.
\begin{table}[H]
	\caption{Results of different weight schemes with corresponding $\beta$ }
	\centering
	\begin{tabular}{lll}
		\toprule
		Reweight Strategy           & mIoU          & $\beta$ \\
		\midrule
		baseline                    & 77.52\%       & 1.1 \\
		signal-reweight             & 77.96\%       & 1.3 \\
		neighbor-reweight           & 78.40\%       & 0.25 \\
		sqrt-reweight               & 78.74\%       & 1.5 \\
		\bottomrule
	\end{tabular}
\end{table}
The weighting scheme of the square root value of inverse frequency has the best performance, as shown in the table below.
The value of $\beta$ for different weight schemes selected by cross-validation is also shown and we can see that the best $\beta$ for neighbor-reweight is much smaller than the others due to the absurdly large weights on $n_0$ and $n_1$. This indicates the necessity of using the sqrt-reweight scheme.

\begin{figure}[h]
	\label{fig:nstat}
	\centering
	\subfigure[Affinity accuracy of sqrt-reweight]{
		\includegraphics[width=0.3\linewidth]{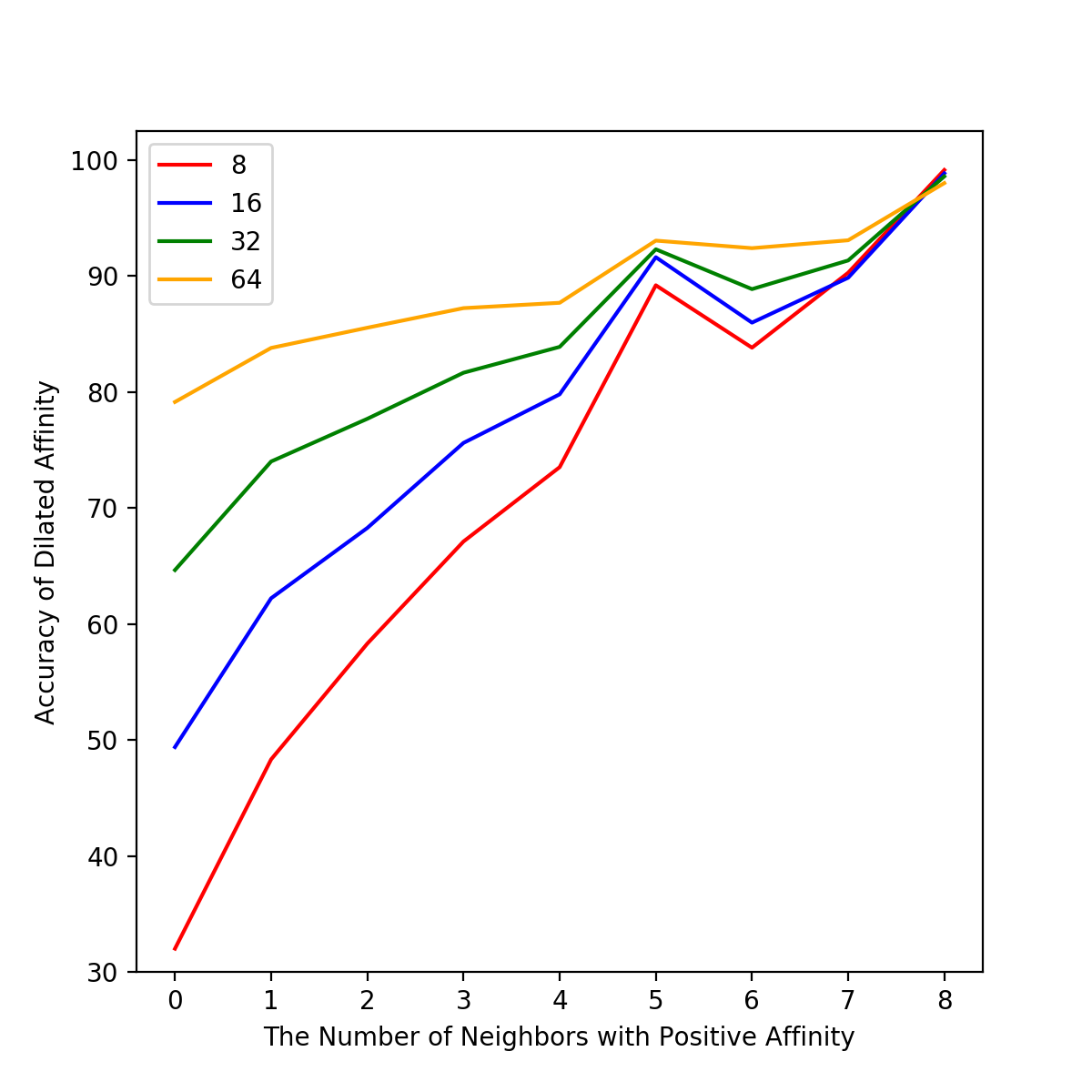}}
	\subfigure[Affinity accuracy of neighbor-reweight]{
		\includegraphics[width=0.3\linewidth]{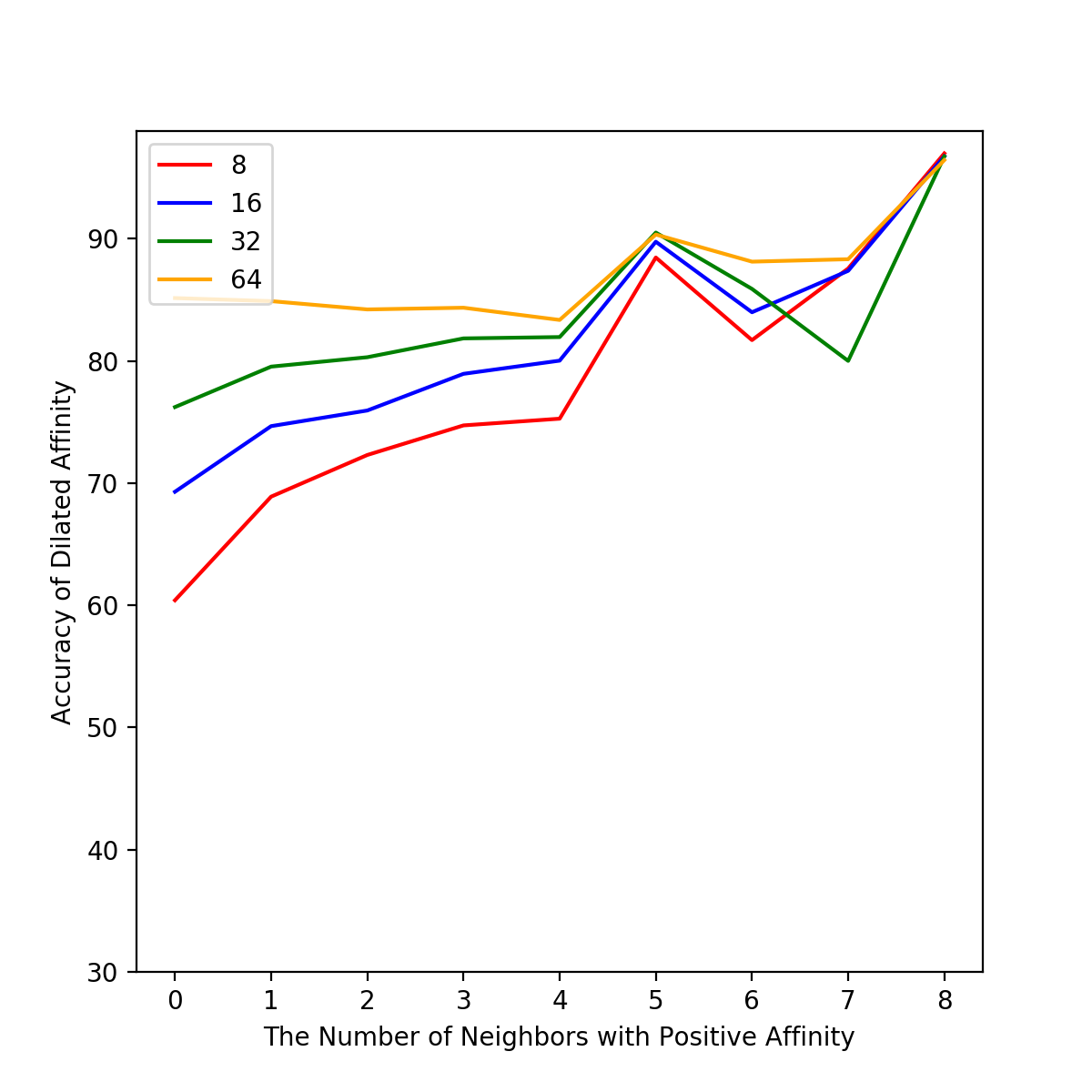}}
	\subfigure[Average affinity accuracy of different weighting schemes]{
		\includegraphics[width=0.3\linewidth]{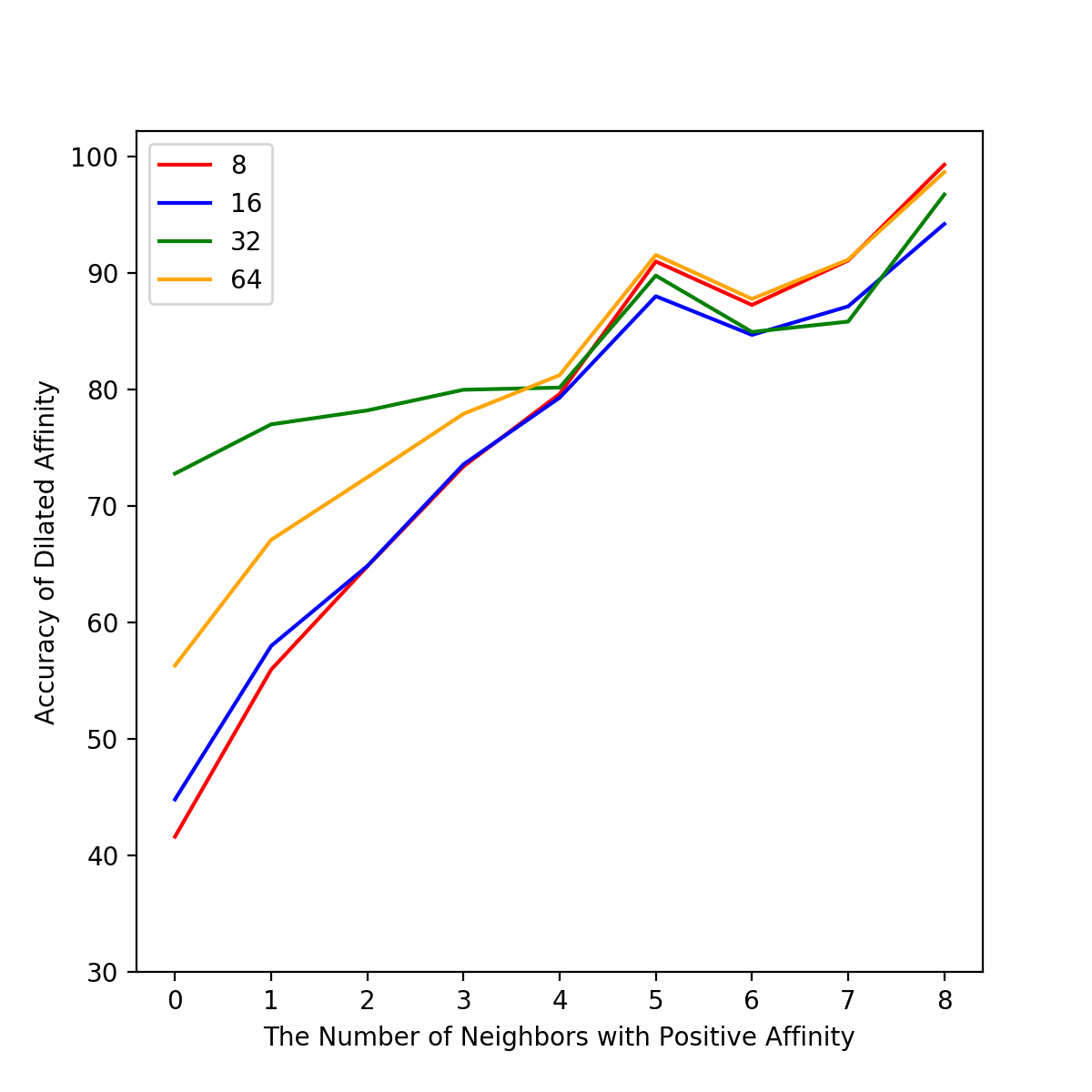}}
	\caption{The accuracy of affinity prediction for $n_0$ to $n_8$ when using sqrt-reweight and neightbor-reweight are shown in (a) and (b). Different lines represent the rates in $\{8,16,32,64\}$. (b) The accuracy of affinity prediction averaged across the rates in $\{8,16,32,64\}$. Different lines represent different weighting schemes.}
	\label{fig:affinity_accu}
\end{figure}

Fig.~\ref{fig:affinity_accu} shows the accuracy of dilated affinity with respect to different weighting schemes and dilation rates.
The accuracies of affinity, especially those of $n_5$ to $n_8$, is important for our affinity propagation process.
For $n_0$ to $n3$, neighbor-reweight has the best performance, while for $n_4$ to $n_8$, sqrt-reweight and baseline achieve a better performance.

We investigate dilated affinity of various dilation rates with the sqrt-reweight scheme. Following the discussion in Sec~\ref{methods}, we explore dilation rates of three aspect ratios, \eg 1:1, 1:2 and 2:1. The first column of the following table denotes the rates. In the brackets, a scalar like $8$ is short for $(8,8)$, while a tuple like $(12,24)$ represents rates of $(12,24)$ as well as $(24,12)$.

\begin{table}[H]
	\caption{Results with different $R$}
	\centering
	\begin{tabular}{lll}
		\toprule
		$R$                        & mIoU       & $\beta$ \\
		\midrule
		$\{8,16\}$                       & 78.96\%    & 1.5     \\
		$\{8,16,32\}$                    & 78.81\%    & 1.2     \\
		$\{8,16,32,64\}$                 & 78.69\%    & 1.6     \\
		$\{8,(12,24),16)\}$              & 78.62\%    & 1.2     \\
		$\{8,(12,24),16,(32,64),48\}$    & 77.62\%    & 0.9     \\
		\bottomrule
	\end{tabular}
\end{table}

% post process
We use the results achieved by the dilated affinity of $\{8,16\}$ and $\{8,(12,24),16\}$ to test our affinity propagation process. We also provide the outcome when we use the ground truth of dilated affinity to refine segmentation results as a supplement.
$\lambda$ is set to 6 according to cross-validation.

\begin{table}[h]
	\caption{Affinity propagation with $R=\{8,16\}$}
	\label{sample-tabl11e}
	\centering
	\begin{tabular}{llllll}
		\toprule
		Iteration times                  & 0       & 1       & 3       & 6       & 10       \\
		\midrule
		Refine with predicted affinity   & 78.96\% & 79.03\% & 79.07\% & 79.10\% & 79.15\%  \\
		Refine with ground truth         & 81.16\% & 81.58\% & 81.94\% & 82.22\% & 82.53\%  \\
		\bottomrule
	\end{tabular}
\end{table}
\begin{table}[h]
	\caption{Affinity propagation with $R=\{8,(12,24),16\}$}
	\label{sample-tabl11e}
	\centering
	\begin{tabular}{llllll}
		\toprule
		Iteration times                  & 0       & 1       & 3       & 6       & 10       \\
		\midrule
		Refine with predicted affinity   & 78.62\% & 78.91\% & 79.09\% & 79.17\% & 79.21\%  \\
		Refine with ground truth         & 82.33\% & 82.94\% & 83.10\% & 83.37\% & 83.59\%  \\
		\bottomrule
	\end{tabular}
\end{table}

Experiments show that dilated affinity with small rates is in favor of the joint training procedure but less effective when used in the affinity propagation stage, while large rates dilated affinity is useful in the propagation stage but may interfere the learning of semantic segmentation. A good practice is based on the tradeoff between these two factors.

\subsection{Experiments on PASCAL VOC 2012}
The best result of our method uses $R=\{8,(12,24),16\}$, $\beta=1.2$, and the sqrt-reweight scheme. We compare it with other methods focusing on the utilization of structural information. Experiment settings of different methods are consistent with the best case in their papers. The mIoU score of our implemented DeepLabv3+ is 1.42\% lower than the one reported in the original paper~\cite{deeplab3plus}.

\begin{table}[H]
	\caption{Results on PASCAL VOC 2012 val set}
	\label{sample-tabl11e}
	\centering
	\begin{tabular}{ll}
		\toprule
		Methods                                     & mIoU  \\
		\midrule
		DeepLabv3+                                  & 77.93\%      \\
		DeepLabv3+ AAF~\cite{aaf}                   & 78.02\%       \\
		DeepLabv3+ CRF~\cite{crf}                   & 78.63\%       \\
		DeepLabv3+ Boundary Reweight~\cite{unet}    & 77.65\%       \\
		DeepLabv3+ Dilated Affinity                 & \textbf{79.21\% }     \\
		\bottomrule
	\end{tabular}
\end{table}

\subsection{Experiments on Cityscapes}
When evaluating the proposed algorithm on Cityscapes dataset, we adopt the best setting on PASCAL VOC, which is $R=\{8,(12,24),16\}$, $\beta=1.2$ and 10 times iterations in affinity propagation. We did not do an exhausting search on these hyperparameters as this part is only to show the applicability.

\begin{table}[h]
	\caption{Results on Cityscapes val set.}
	\centering
	\begin{tabular}{ll}
		\toprule
		Methods             & mIoU      \\
		\midrule
		DeepLabv3+          & 77.15\%      \\
		DeepLabv3+ Dilated Affinity       & 78.70\%     \\
		\bottomrule
	\end{tabular}
	
\end{table}

\section{Conclusion}
\label{conclusion}

Our proposed extra learning of dilated affinity information can consistently improve the current state-of-the-art method with minor extra cost. Dilated affinity can assist the original task of semantic segmentation from two aspects. First, joint training with dilated affinity helps the learning process of semantic segmentation. Second, segmentation results can be refined with a fast affinity propagation post-processing, which exploits the extra information generated by the network. Different choices of how to learn the extra information are fully explored in our paper and the learned hyperparameters can be extended to other datasets.

\medskip

\small
\bibliographystyle{plain}
\bibliography{egbib}

\end{document}